\documentclass{article} % For LaTeX2e
\usepackage{iclr2023_conference_tinypaper,times}
\usepackage{graphicx}

\usepackage{amsmath,amsfonts,bm}

\def\eqref#1{equation~\ref{#1}}
\def\1{\bm{1}}

\DeclareMathAlphabet{\mathsfit}{\encodingdefault}{\sfdefault}{m}{sl}
\SetMathAlphabet{\mathsfit}{bold}{\encodingdefault}{\sfdefault}{bx}{n}

\usepackage{hyperref}

\usepackage{url}
\usepackage{amsthm}
\usepackage{cleveref}

\theoremstyle{definition}
\newtheorem{defn}{Definition}[section]

\theoremstyle{plain}
\newtheorem{thm}[defn]{Theorem}
\newtheorem{lem}[defn]{Lemma}

\newtheorem{assum}[defn]{Assumption}

\newtheorem{imp}[defn]{Implication}

\theoremstyle{remark}
\newtheorem{rem}[defn]{Remark}

\usepackage{subcaption}
\usepackage{xcolor}
\usepackage{algorithm}
\usepackage{algorithmic}

\title{DFWLayer: Differentiable Frank-Wolfe Optimization Layer}

\author{Zixuan Liu$^1$, Liu Liu$^2$, Xueqian Wang$^1$, Peilin Zhao$^2$ \\
$^1$Tsinghua University, $^2$Tencent AI Lab\\
\texttt{\{zx-liu21@mails,wang.xq@sz\}.tsinghua.edu.cn,} \\
\texttt{\{leonliuliu,masonzhao\}@tencent.com} \\
}

\iclrfinalcopy % Uncomment for camera-ready version, but NOT for submission.
\begin{document}

\maketitle

\begin{abstract}
Differentiable optimization has received a significant amount of attention due to its foundational role in the domain of machine learning based on neural networks. This paper proposes a differentiable layer, named \underline{D}ifferentiable \underline{F}rank-\underline{W}olfe Layer (DFWLayer), by rolling out the Frank-Wolfe method, a well-known optimization algorithm which can solve constrained optimization problems without projections and Hessian matrix computations, thus leading to an efficient way of dealing with large-scale convex optimization problems with norm constraints. Experimental results demonstrate that the DFWLayer not only attains competitive accuracy in solutions and gradients but also consistently adheres to constraints. 
\end{abstract}

\section{Introduction \& Related Work}

% \TODO{DO Layer explicit function not easy}

Recent years have witnessed a variety of combining neural networks and conventional optimization as differentiable optimization layers to integrate expert knowledge into machine learning systems. With objectives and constraints deriving from that knowledge, the output of each differentiable optimization layer is the solution to a specific optimization problem whose parameters are outputs from previous layers \citep{amos2017optnet,agrawal2019cvxpy,sun2022alternating,landry2021differentiable}. Nevertheless, adapting the implicit functions (mapping parameters to solutions) to the training procedure of deep learning architecture is not easy since the explicit differentiable closed-form solutions are not available for most conventional optimization algorithms. On top of directly obtaining differentiable closed-form solutions, one alternative is to recover gradients respect to some parameters after solving for the optimal solution \citep{landry2021differentiable}.

% \TODO{DFWLayer}

In this study, we introduce a novel differentiable unrolling optimization layer designed to enhance the speed of both optimization and backpropagation processes in the presence of norm constraints. Drawing inspiration from the Frank-Wolfe algorithms \citep{frank1956fwalgorithm}, also known as conditional gradient algorithms, we have developed the Differentiable Frank-Wolfe Layer (DFWLayer). This layer is specifically tailored to efficiently handle norm constraints where projection onto the corresponding feasible region is computationally expensive.

In terms of related work, two main categories of recovering gradients for optimal solutions have emerged: differentiating the optimality conditions \citep{amos2017optnet,agrawal2019cvxpy} and rolling out solvers \citep{donti2020dc3}. Furthermore, \citet{sun2022alternating} leverage on alternating direction method of multipliers (ADMM) \citep{boyd2011admm} to differentiate the optimality conditions at each iterative step. However, these existing methods often encounter challenges. They can be computationally intensive or suffer from prolonged convergence times, leading to suboptimal solutions that potentially diminish the overall system performance \citep{bambade2023qplayer}.

\section{Methodology}\label{sec:method}

In this section, we consider such an convex optimization problem with norm constraints, $\min_x f(x;\theta) \ \text{s.t.} \ \lVert w \circ x \rVert \le t$,
where $f: \mathbb{R}^n \rightarrow \mathbb{R}$ is convex and $L$-smooth, and $\mathcal{C}(\theta)$ is the convex feasible region; $x \in \mathbb{R}^n$ is the variable, and $\theta \in \mathbb{R}^m$ are the parameters of the optimization problem; $w \in \mathbb{R}^{n}$ is a weight vector, $t \in \mathbb{R}^+$ is a non-negative constant, and $\circ$ denotes the Hadamard production.

Given the widespread use of $\ell_p$-norm constraints\footnote{As for others, for instance, trace norm can be considered with its dual norm: operator norm.} in various applications, we considering $\ell_p$-norm constraints, $\lVert w \circ x \rVert_p \le t$, so that the linear approximation of the original objective problem, $\arg\min_{s \in \mathcal{C}} \langle\ \nabla_x f(x_{k}), s \rangle$, can solved by $s_{k,i} = -\frac{\alpha t}{w} \cdot {\rm sign} (g_{tw,i}(x_k)) \cdot \lvert g_{tw,i}(x_k) \rvert^{\frac{p}{q}}$,
where $g_{tw}(\cdot)=\frac{t}{w} \circ \nabla_x f(\cdot)$, $i=1,...,n$ stands for the $i^{\rm th}$ entry of the vector, $q$ is a positive constant such that $\frac{1}{p}+\frac{1}{q}=1$, and $\alpha$ is a constant such that $\lVert s_k \rVert_q = 1$. It is noted that, when $p>1$, $s_{k,i}$ is differentiable and can directly integrated into automatic differentiation.

When $p=1$, a differentiable expression of vertex is $\hat{s}_k=-\frac{t}{w} \circ {\rm sign}(g_{tw}(\hat{x}_k)) \circ {\rm softmax}(r(\hat{x}_k)/\tau)$ with probability approximation, where $r(\cdot) = \lvert g_{tw}(\cdot)\rvert$, and $\hat{p}_k(\tau)=\hat{p}(\cdot \mid r(\hat{x}_k);\tau)$ for simplicity, and $\tau$ is the temperature. According to \Cref{thm:gap}, an annealing temperature is designed as $\tau_k = 2^{-k//T}$, which decreases per $T$ steps, to obtain reliable solutions and gradients at the same time. 

\begin{thm}\label{thm:gap}
    Let $f: \mathbb{R}^n \rightarrow \mathbb{R}$ be a L-smooth convex function on a convex region $\mathcal{C}$ with diameter $M$ and $x^\ast=\arg\min_{x\in\mathcal{C}} f(x)$. Under \Cref{assum:dis}, the suboptimality gap of DFWLayer for $\ell_1$ norm constraints is bounded by
    \begin{align}
        h(\hat{x}_k) = f(\hat{x}_k) - f(x^\ast) \le \delta(\tau_k) + \frac{5LM^2}{2k+4}.\label{eq:thm}
    \end{align}
\end{thm}

The proof of \Cref{thm:gap} is left to \Cref{sec:thm}. Therefore, through the differentiable step size $\gamma_k = \min\{\frac{\langle \nabla_x f(\hat{x}_{k}), \hat{x}_{k}-\hat{s}_{k}\rangle}{L\rVert \hat{x}_{k}-\hat{s}_{k}\rVert^2}, 1\}$ and the Frank-Wolfe update $\hat{x}_{k+1}=(1-\gamma_{k})\hat{x}_{k}+\gamma_{k} \hat{s}_{k}$\footnote{$\hat{x}_k = x_k$ and $\hat{s}_k = s_k$ for $p>1$ without probability approximation.}, the automatic differentiation can be involved to obtain the derivatives $\frac{\partial \hat{x}^\ast}{\partial \theta}$ for general $\ell_p$-norm constraints.

\section{Experimental Results}

In this section, we evaluate DFWLayer with several experiments to validate its performance to solve  different-scale quadratic programs regarding the efficiency and accuracy, compared with state-of-the-art methods. Besides, the results for robotics tasks are left to \Cref{sec:rob}. We choose CvxpyLayer~\citep{agrawal2019cvxpy} and Alt-Diff~\citep{sun2022alternating} as the baselines. The code is available in \url{https://github.com/Panda-Shawn/DFWLayer}.

As is shown in \Cref{tab:time}, DFWLayer performs much more efficiently than other two baselines and the speed gap becomes larger as the problem scale, namely the dimension of variable, increases. Meanwhile, DFWLayer also has a competitive accuracy and results can be found in \Cref{sec:large_scale}.
 
\begin{table*}[htbp]
    \centering
    \caption{Comparison of running time (s) between different-scale optimization problems. The average and standard deviation are obtained over 5 trials. Lower values are better for the running time.}
    \begin{tabular}{c|c|c|c}
        \hline
         & Small & Medium & Large \\
        \hline
        Dimension of variable & 500 & 1000 & 2000 \\
        \hline
        CvxpyLayer &
7.68 $\pm$ 0.07 & 59.02 $\pm$ 0.16 & 481.78 $\pm$ 2.54 \\
        Alt-Diff &
1.05 $\pm$ 0.14 & 3.84 $\pm$ 0.12 & 22.51 $\pm$ 0.13 \\
        DFWLayer &
\textbf{0.18 $\pm$ 0.01} & \textbf{0.21 $\pm$ 0.01} & \textbf{0.31 $\pm$ 0.02} \\
        \hline
    
    \end{tabular}
    \label{tab:time}
\end{table*}

\section{Conclusions}
In this paper, we have proposed DFWLayer for solving convex optimization problems with $L$-smooth functions and norm constraints in an efficient way. Naturally derived from the Frank-Wolfe, DFWLayer accelerates to obtain solutions and gradients based on first-order optimization methods which avoid projections and Hessian matrix computations. Especially for $\ell_1$-norm constraints, DFWLayer modifies non-differentiable operators with probabilistic approximation  so that gradients can be efficiently computed through the unrolling sequence with automatic differentiation. Also, an annealing temperature is designed to guarantee the quality of both solutions and gradients. In future work, we plan to address a significant limitation of the DFWLayer: its current restriction to handling only norm constraints. We aim to explore and adapt the DFWLayer for a broader range of assumptions, thereby expanding its applicability in more diverse computational scenarios.

\subsubsection*{URM Statement}

The authors acknowledge that at least one key author of this work meets the URM criteria of ICLR 2024 Tiny Papers Track.

\bibliography{iclr2023_conference_tinypaper}
\bibliographystyle{iclr2023_conference_tinypaper}
\newpage
\appendix
\section{Appendix}
\subsection{Algorithm Details}

The DFWLayer for $\ell_1$-norm constraints is presented
in \Cref{alg:dfw}.

\begin{algorithm}[h]
    \caption{Differentiable Frank-Wolfe Layer}\label{alg:dfw}
    \textbf{Input}: Parameters $\theta$ from the previous layers. \\
    \textbf{Parameter}: Parameters $w$ and $t$.\\
    \textbf{Output}: The optimal solutions to the problem:  $\min_x f(x;\theta) \ \text{s.t.} \ \lVert w \circ x \rVert_1 \le t$.\\
    \begin{algorithmic}[1]

        \STATE \textbf{Forward:}
        % \[\]
        \STATE Let $x_0 \in \mathcal{C}$ and $\hat{x}_0 = x_0$
        \FOR{$k=0,...,K$}
        \STATE Compute $\hat{s}_{k}$ using \Cref{eq:probvertexhat}.
        \STATE Obtain $\gamma_k = \min\{\frac{\langle \nabla_x f(\hat{x}_{k}), \hat{x}_{k}-\hat{s}_{k}\rangle}{L\rVert \hat{x}_{k}-\hat{s}_{k}\rVert^2}, 1\}$.
        \STATE Update $\hat{x}_{k+1}$ by $\hat{x}_{k+1}=(1-\gamma_k)\hat{x}_{k}+\gamma_k \hat{s}_{k}$.
        \ENDFOR 
        \[\]
        \STATE \textbf{Backward:}
        % \[\]
        \STATE Obtain $\frac{\partial \hat{x}^\ast}{\partial \theta}$ with automatic differentiation.
    \end{algorithmic}
\end{algorithm}

\subsection{Theoretical Results}\label{sec:thm}

In this section, we provides convergence analysis for DFWLayer dealing with $\ell_1$-norm constraints. The exploration begins on a modest assumption, keeping the analysis approachable and generalizable.

Inspired by \citep{dalle2022prob} turning combinatorial optimization layers into probabilistic layers, searching vertex of the feasible region can be regarded as the expectation from $\mathcal{S}=\{s^1,...,s^{2n} \}$, which denotes the vertices of $\mathcal{C}$, over a probabilistic distribution $p(\cdot \mid \nabla_x f(x_k))$,
\begin{align} \label{eq:probvertex}
    s_k = \mathbb{E}_{s\sim p(\cdot \mid \nabla_x f(x_k)) } [S] =\sum_{s \in \mathcal{s}} s \cdot p(s \mid \nabla_x f(x_k)).
\end{align}

Intuitively, the probabilistic distribution can be chosen as the Dirac mass $p(\cdot \mid \nabla_x f(x_k)) = \delta_{s^\ast}(s)$ with $s^\ast=\arg\min_{s\in\mathcal{S}} \langle \nabla_x f(x_k),s \rangle$, though sharing the lacks of differentiability with $\arg\max$. Thus, we aim to find a differentiable $\hat{p}(\cdot \mid \nabla_x f(x_k))$ so that the Jacobian matrix of $s_k$ is tractable, and the Boltzmann distribution is an appropriate candidate, leading to $\hat{p}(e_i \mid \nabla_x f(x_k)) \propto e^{\nabla_x f(x_k)^T e_i}=e^{ \nabla_{x,i} f(x_k)}$. Thus, we can derive \Cref{eq:probvertex} as
\begin{align} \label{eq:probvertexhat}
    \hat{s}_k &= \mathbb{E}_{s\sim \hat{p}_k(\tau) } [s] \nonumber \\ &=-\frac{t}{w} \circ {\rm sign}(g_{tw}(\hat{x}_k)) \circ \sum_{i} e_i \cdot \frac{e^{r_i(\hat{x}_k)/\tau}}{\sum_{i} e^{r_i(\hat{x}_k)/\tau}} \nonumber \\ &=-\frac{t}{w} \circ {\rm sign}(g_{tw}(\hat{x}_k)) \circ {\rm softmax}(r(\hat{x}_k)/\tau),
\end{align}

% \begin{assum}\label{assum:dis}
%     The approximating distribution $\hat{p}$ and the target distribution $p$ are similar to each other such that:
%     \begin{align}
%         \langle \nabla_x f(\hat{x}_k), \Bar{s}_k\rangle \le \langle \nabla_x f(\hat{x}_k), \hat{s}_k\rangle \le \langle \nabla_x f(\hat{x}_k), x^\ast\rangle,
%     \end{align}
%     where $\Bar{s}_k = \mathbb{E}_{s\sim p(\cdot \mid \nabla_x f(\hat{x}_k)) } [S]$, and $\hat{s}_k = \mathbb{E}_{s\sim \hat{p}(\cdot \mid \nabla_x f(\hat{x}_k)) } [S]$, and $x^\ast = \arg\min_{x\in \mathcal{C}}f(x)$.
% \end{assum}

\begin{assum}\label{assum:dis}
    The Maximum Mean Discrepancy (MMD) between the approximating distribution $\hat{p}$ and the target distribution $p$ is upper bounded by
    \begin{align}
        &{\rm MMD}(\Psi, \hat{p}_k(\tau), p_k)\nonumber\\
        &=\sup_{\psi\in\Psi}\mathbb{E}_{\hat{p}_k(\tau)}[\psi(s)]-\mathbb{E}_{p_k} [\psi(s)]\le \delta(\tau), \
        \text{for all k},
    \end{align}
    where $\delta(\tau)$ is a positive constant related to temperature $\tau$.
\end{assum}

Considering the temperature $\tau$ of the ${\rm softmax}$ function, \Cref{assum:dis} is reasonably modest because the ${\rm softmax}$ function can always approach the ${\rm hardmax}$ function with a small temperature and thus $\delta(\tau)\rightarrow0$ when $\tau\rightarrow0$.

\begin{imp}\label{cor:mmd}
    Under \Cref{assum:dis}, for all $p_k$, there exists a temperature $\tau_k$ for $\hat{p}_k(\tau_k)$ such that $\delta(\tau_k)\le O(\frac{1}{k})$.
\end{imp}

The relationship between the ${\rm hardmax}$ and the ${\rm softmax}$ function makes \Cref{cor:mmd} directly follow \Cref{assum:dis}. With the implication, we can further understand \Cref{thm:gap}.

The theorem has shown that the suboptimality is bounded by the approximation gap and the origin gap. According to \Cref{cor:mmd}, if $\delta(\tau_k) \le O(\frac{1}{k})$ through a certain $\tau_k$, we can say DFWLayer converges at a sublinear rate. In \cref{sec:method}, we discuss an annealing temperature $\tau_k = 2^{-k//T}$ which decreases per $T$ steps, and validate its improvement compared to constant temperatures.

As for the proof of \Cref{thm:gap}, we can first split it into two terms,
\begin{align} \label{eq:split}
    h(\hat{x}_k) &= f(\hat{x}_k) - f(x^\ast) \nonumber \\
    &= \underbrace{f(\hat{x}_k) - f(x_k)}_{\text{approximation gap}} + \underbrace{f(x_k) - f(x^\ast)}_{\text{origin gap}}.
\end{align}

The original gap of the Frank-Wolfe with the short path rule and its proof are provided by \citet{braun2022conditional}, we restate it as the following lemma.

\begin{lem}\label{lem}
    Let $f: \mathbb{R}^n \rightarrow \mathbb{R}$ be a L-smooth convex function on a convex region $\mathcal{C}$ with diameter $M$. The Frank-Wolfe with short path rule converges as follows:
    \begin{align}
        h(x_k) = f(x_k) - f(x^\ast)\le \frac{2LM^2}{k+3}\le \frac{2LM^2}{k+2}.
    \end{align}
\end{lem}

Then, we need to present a useful remark, which is used for the proof of \Cref{lem} in \citep{braun2022conditional} and the following proof of \Cref{thm:gap}.

\begin{rem} \label{rem}
    In essence, the proof of \Cref{thm:gap} uses the modified agnostic step size $\gamma_k = \frac{2}{k+3}$, instead of the standard agnostic step size $\gamma_k = \frac{2}{k+2}$, because the convergence rate for the short path rule dominates the modified agnostic step size.
\end{rem}

With \Cref{lem} and \Cref{rem}, we now start to prove \Cref{thm:gap}.
\begin{proof}[Proof of \Cref{thm:gap}]
    The original solutions $x_k$ can be recursively expressed as the combination of initial $x_0$ and vertex $s_k$ at each iteration,
\begin{align} \label{eq:ori_sol}
    x_k &= (1-\gamma) x_{k-1} + \gamma s_{k-1} \nonumber \\
    &= (1-\gamma) ((1-\gamma) x_{k-2} + \gamma s_{k-2}) + \gamma s_{k-1} \nonumber \\
    &= (1-\gamma)^k x_0 + \gamma \sum_o^{k-1} (1-\gamma)^{k-1-i} s_i.
\end{align}
Similar with the original solutions, the solutions obtained by DFWLayer can also be derived as follows:
\begin{align}  \label{eq:dfw_sol}
    \hat{x}_k = (1-\gamma)^k x_0 + \gamma \sum_o^{k-1} (1-\gamma)^{k-1-i} \hat{s}_i.
\end{align}
And thus we subtract \Cref{eq:ori_sol} from \Cref{eq:dfw_sol} and obtain
\begin{align} \label{eq:sub}
    \hat{x}_k -x_k = \gamma \sum_o^{k-1} (1-\gamma)^{k-1-i} (\hat{s}_i - s_i).
\end{align}

Considering the smoothness of $f$ and \Cref{eq:sub}, the approximation can be derived as
\begin{align} 
    f(\hat{x}_{k}) - f(x_k) &\le \langle \nabla_x f(x_k), \hat{x}_k - x_k \rangle + \frac{L}{2} \lVert \hat{x}_k - x_k \rVert^2 \nonumber \\
    &= \gamma \sum_0^{k-1} (1-\gamma)^{k-1-i} \nabla_x f(x_k)^T (\hat{s}_i - s_i) + \frac{L}{2} \lVert \gamma \sum_0^{k-1} (1-\gamma)^{k-1-i} (\hat{s}_i - s_i) \rVert^2 \nonumber \\
    &\le \gamma \sum_0^{k-1} (1-\gamma)^{k-1-i} (\mathbb{E}_{s\sim \hat{p}_i} [\nabla_x f(x_k)^Ts] - \mathbb{E}_{s\sim p_i} [\nabla_x f(x_k)^Ts]) \nonumber \\
    &+ \frac{L}{2} \gamma^2 \sum_0^{k-1} (1-\gamma)^{2(k-1-i)} \lVert \hat{s}_k - s_k \rVert^2 \nonumber \\
     &\le \gamma \sum_0^{k-1} (1-\gamma)^{k-1-i} \delta + \frac{L}{2} \gamma^2 \sum_0^{k-1} (1-\gamma)^{2(k-1-i)} M^2 \nonumber \\
     &= (1-(1-\gamma)^k) \delta + \frac{LM^2}{2} \frac{\gamma}{2-\gamma} (1-(1-\gamma)^{2k}) \nonumber \\
     &\le \delta + \frac{LM^2}{2k+4}.
\end{align}
Here, we use the triangle inequality in the first inequality. Then, we bound the difference of expectations by $\delta$ under \Cref{assum:dis} and the distance between vertices by region diameter $M$ in the second inequality. The last equality is obtained by choosing $\gamma_k = \frac{2}{k+3}$ via heuristic in \Cref{rem}.

Therefore, substituting the approximation gap from \Cref{lem} and the origin gap in \Cref{eq:split} yields \Cref{eq:thm} as claimed.
\end{proof}

\subsection{Experimental Details}\label{sec:exp}

All the experiments were implemented on an Intel(R) Xeon(R) Platinum 8255C CPU @ 2.50GHz with 40 GB of memory and a Tesla T4 GPU.

\subsubsection{Different-Scale Optimization Problems} \label{sec:large_scale}

This section details the experiments in implementing our method to tackle quadratic programs of varying scales. We clarify the setup and discuss how our method performs in terms of both efficiency and accuracy, as well as the ${\rm softmax}$ temperature.

\paragraph{Efficiency.}

With objective $f(x; q)=\frac{1}{2}x^TPx+q^Tx$ which chooses $q$ as the parameter requiring gradients and constraints $\lVert w\circ x\rVert \le t$, all the parameters $P, q, w, t$ are generated randomly with symmetric matrix $P \succeq 0$ and $t \geq 0$. The tolerance $\epsilon = \frac{\lvert f(x_{k+1})-f(x_k)\rvert}{\lvert f(x_k)\rvert}$ is used to terminate DFWLayer instead of a maximum iteration number. For all the methods, the tolerance is set as $\epsilon = 1e-4$. We executed all the experiments 5 times and reported the average running time with standard deviation in \Cref{tab:time}. It is shown that DFWLayer performs much more efficiently than other two baselines and the speed gap becomes larger as the problem scale increases.

\begin{table*}[htbp]
    \centering
    \caption{Comparison of accuracy w.r.t. solutions and gradients for medium-scale problems. The average and standard deviation are obtained over 5 trials. Higher values are better for the gradients similarity, while lower values are better for others. Constraints violation is shown in \textcolor{red}{red}. Gradients similarity and solutions distance are computed with those obtained by CvxpyLayer as the reference.}
    \begin{tabular}{c|c|c|c|c}
        \hline
         & Gradients Sim. & Solutions Dist. & Mean Violation & Max Violation \\
        \hline
        CvxpyLayer & 1.000 $\pm$ 0.000 & 0.000 $\pm$ 0.000 & \color{blue}{0.000} & \color{blue}{0.000} \\
        Alt-Diff &
\textbf{0.980 $\pm$ 0.040} & \textbf{0.001 $\pm$ 0.002} & \color{red}{0.058} & \color{red}{1.000} \\
        DFWLayer & \textbf{0.980 $\pm$ 0.021} & 0.002 $\pm$ 0.001 & \color{blue}{0.000} & \color{blue}{0.000} \\
        
        \hline
    
    \end{tabular}
    \label{tab:accu}
\end{table*}

\begin{table*}[htbp]
    \centering
    \caption{Comparison of accuracy w.r.t. solutions and gradients for small-scale problems. The average and standard deviation are obtained over 5 trials. Higher values are better for the gradients similarity, while lower values are better for others. Constraints violation is shown in \textcolor{red}{red}. Gradients similarity and solutions distance are computed with those obtained by CvxpyLayer as the reference.}
    \begin{tabular}{c|c|c|c|c}
        \hline
         & Gradients Sim. & Solutions Dist. & Mean Violation & Max Violation \\
        \hline
        CvxpyLayer & 1.000 $\pm$ 0.000 & 0.000 $\pm$ 0.000 & \color{blue}{0.000} & \color{blue}{0.000} \\
        Alt-Diff & 0.975 $\pm$ 0.049 & \textbf{0.002 $\pm$ 0.003} & \color{red}{0.063} & \color{red}{1.000} \\
        DFWLayer & \textbf{0.977 $\pm$ 0.025} & \textbf{0.002 $\pm$ 0.001} & \color{blue}{0.000} & \color{blue}{0.000} \\
        
        \hline
    
    \end{tabular}
    \label{tab:accu_small}
\end{table*}

\begin{table*}[htbp]
    \centering
    \caption{Comparison of accuracy w.r.t. solutions and gradients for large-scale problems. The average and standard deviation are obtained over 5 trials. Higher values are better for the gradients similarity, while lower values are better for others. Constraints violation is shown in \textcolor{red}{red}. Gradients similarity and solutions distance are computed with those obtained by CvxpyLayer as the reference.}
    \begin{tabular}{c|c|c|c|c}
        \hline
         & Gradients Sim. & Solutions Dist. & Mean Violation & Max Violation \\
        \hline
        CvxpyLayer & 1.000 $\pm$ 0.000 & 0.000 $\pm$ 0.000 & \color{blue}{0.000} & \color{blue}{0.000} \\
        Alt-Diff & \textbf{0.978 $\pm$ 0.044} & \textbf{0.001 $\pm$ 0.001} & \color{red}{0.061} & \color{red}{1.000} \\
        DFWLayer & \textbf{0.978 $\pm$ 0.023} & \textbf{0.001 $\pm$ 0.001} & \color{blue}{0.000} & \color{blue}{0.000} \\
        
        \hline
    
    \end{tabular}
    \label{tab:accu_large}
\end{table*}

\paragraph{Accuracy.} As for accuracy, we report the cosine similarity of gradients and the Euclidean distance of solutions with CvxpyLayer as the reference for its stable performance. Also, constraints violation is a significant index for constrained optimization expecially when these constraints stand for safety. Thus, we show the results for medium problem scale in \Cref{tab:accu}. It should be noted that we state the maximum violation instead of standard deviation for its widely use in practice. It can be seen from \Cref{tab:accu} that the solutions provided by Alt-Diff violate the constraints, although it has the higher accuracy of solutions than DFWLayer. The constraints violation of Alt-Diff results from insufficient iterations and a lack of projection to ensure the feasibility for its solutions, while our method enforces the feasibility without projection and thus solves the optimization problems in much less time with an acceptable accuracy margin.

\paragraph{Softmax Temperature.} The ${\rm softmax}$ temperature is a crucial hyperparameter for DFWLayer, because it can assist to approximate the target distribution, which is required by \Cref{assum:dis}. The annealing schedule $\tau_k = 2^{-k//30}$ is used for the proposed method, so that the distance between the target and the approximating distributions can get closer during the iteration. In order to give a clear explanation of this choice, we plot the curve of cosine similarity and Euclidean distance compared with different constant temperatures.

\begin{figure}[ht]
\centering

\begin{subfigure}[b]{0.48\columnwidth}
\centering  
\includegraphics[width=0.9\textwidth]{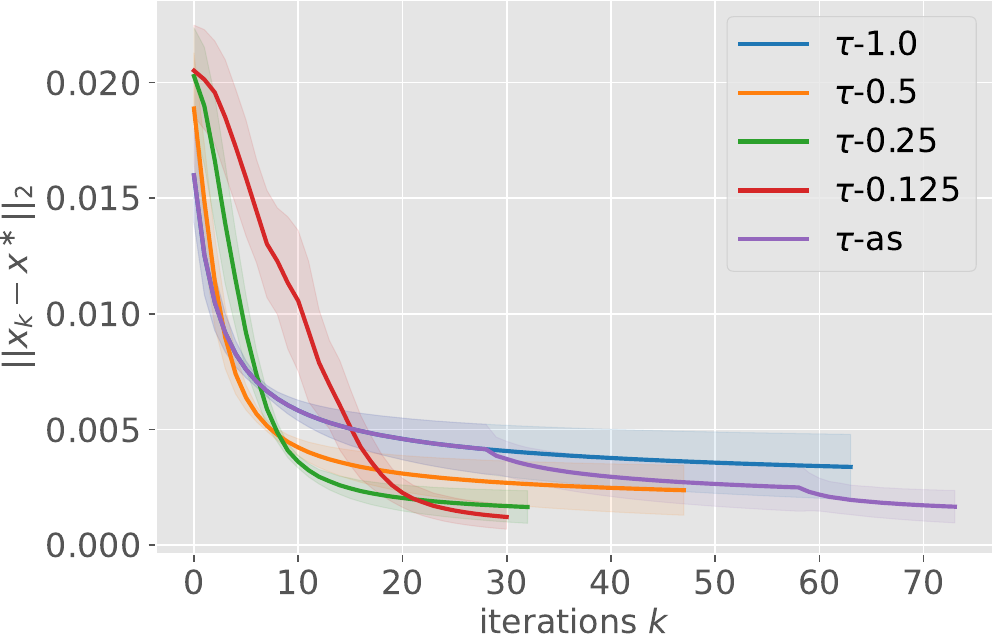} 
\caption{Solutions distance.}
\label{fig:edis}
\end{subfigure}
\begin{subfigure}[b]{0.48\columnwidth}
\centering
\includegraphics[width=0.9\textwidth]{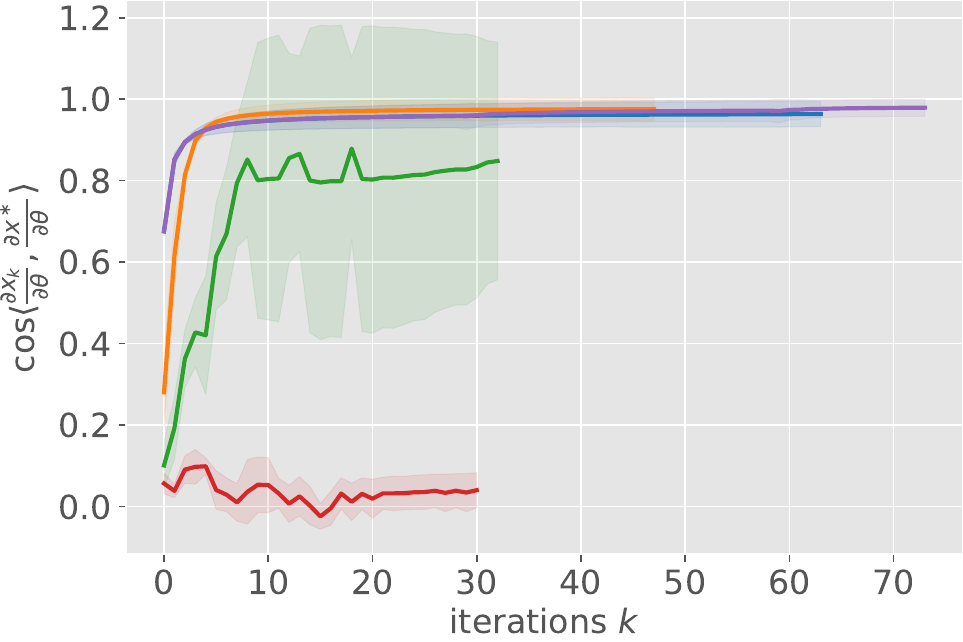} 
\caption{Gradients similarity.}
\label{fig:cos}
\end{subfigure}%
\caption{Gradients and solutions distance between CvxpyLayer and DFWLayer with different temperatures for medium-scale problems. All the curves terminate for tolerance $\epsilon = 1e-4$. The shaded area for all the curves stands for standard deviation over 5 trials and $x^\ast$ stands for solutions obtained by CvxpyLayer.}\label{fig:tmp}
\end{figure}

\begin{figure*}[ht]
\centering
\begin{subfigure}[b]{0.3\columnwidth}
\centering  
\includegraphics[width=1.0\textwidth]{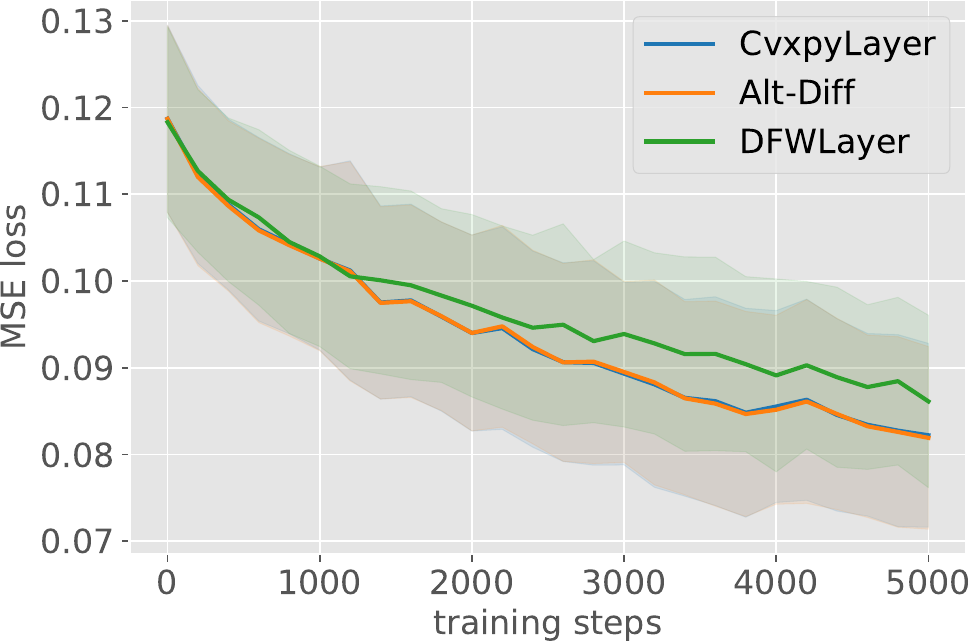} 
\caption{MSE loss.}
\label{fig:roloss}
\end{subfigure}
\begin{subfigure}[b]{0.3\columnwidth}
\centering
\includegraphics[width=1.0\textwidth]{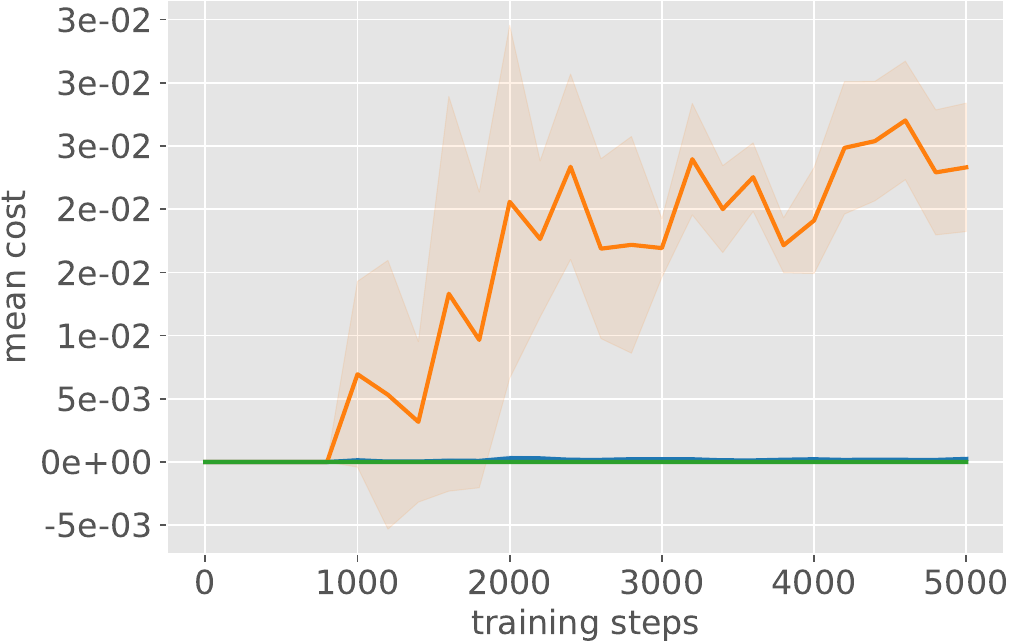} 
\caption{Mean violation.}
\label{fig:rocost}
\end{subfigure}
% \begin{subfigure}[b]{1.0\columnwidth}
% \centering  
% \includegraphics[width=0.9\textwidth]{edis_tau.pdf} 
% \caption{Solutions distance.}
% \label{fig:vio}
% \end{subfigure}
\begin{subfigure}[b]{0.3\columnwidth}
\centering
\includegraphics[width=1.0\textwidth]{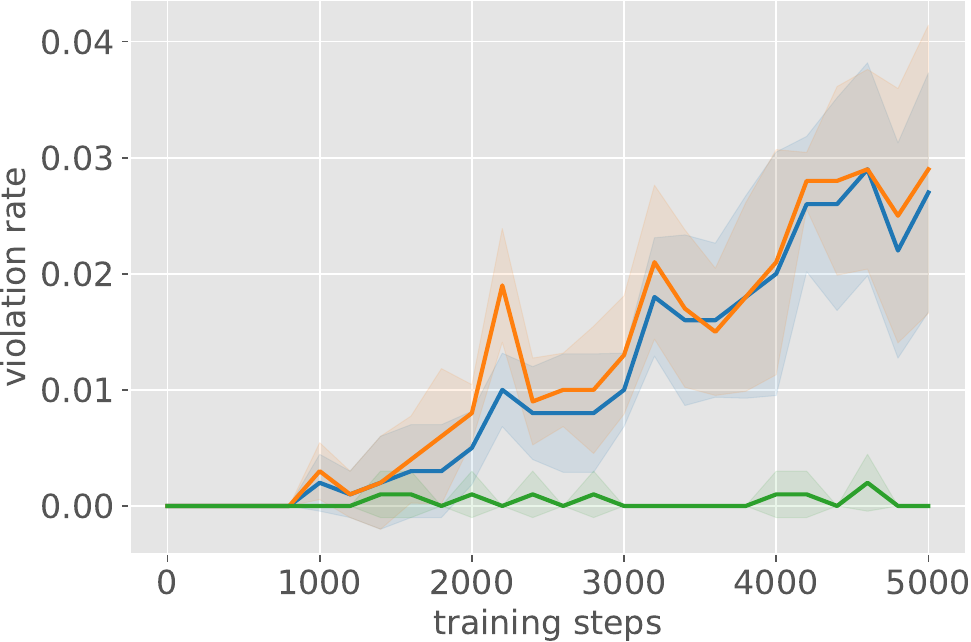} 
\caption{Violation rate.}
\label{fig:rovio}
\end{subfigure}
% \caption{MSE loss, mean violation and violation rate for R+O03. Mean violation is computed over violated samples and violation rate is the ratio of violated samples to all samples in the testing set. The shaded area for all the curves stands for standard deviation over 5 trials.}\label{fig:ro03}
% \end{figure*}

% \begin{figure*}[ht]
% \centering
\begin{subfigure}[b]{0.3\columnwidth}
\centering  
\includegraphics[width=1.0\textwidth]{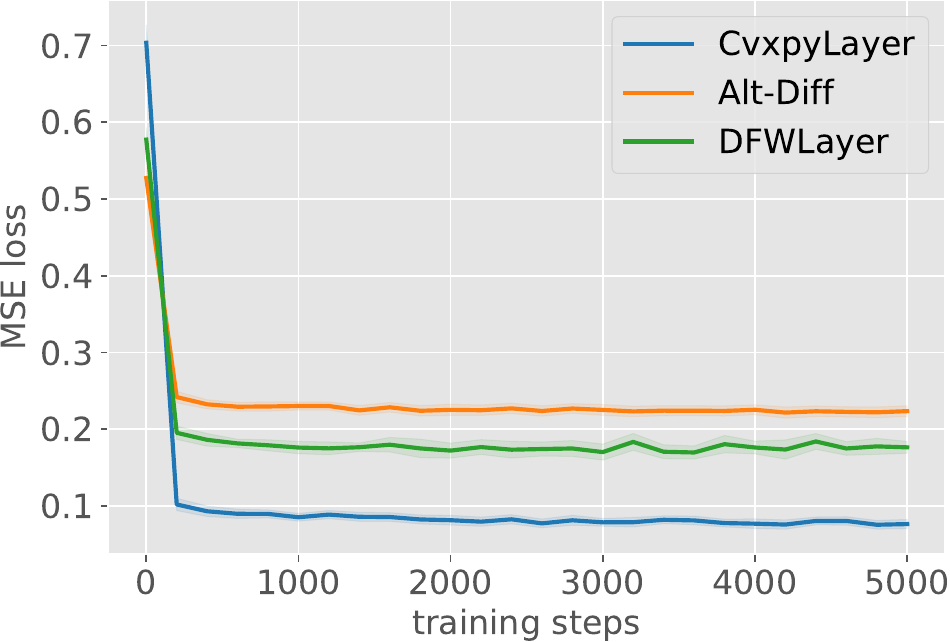} 
\caption{MSE loss.}
\label{fig:hcoloss}
\end{subfigure}
\begin{subfigure}[b]{0.3\columnwidth}
\centering
\includegraphics[width=1.0\textwidth]{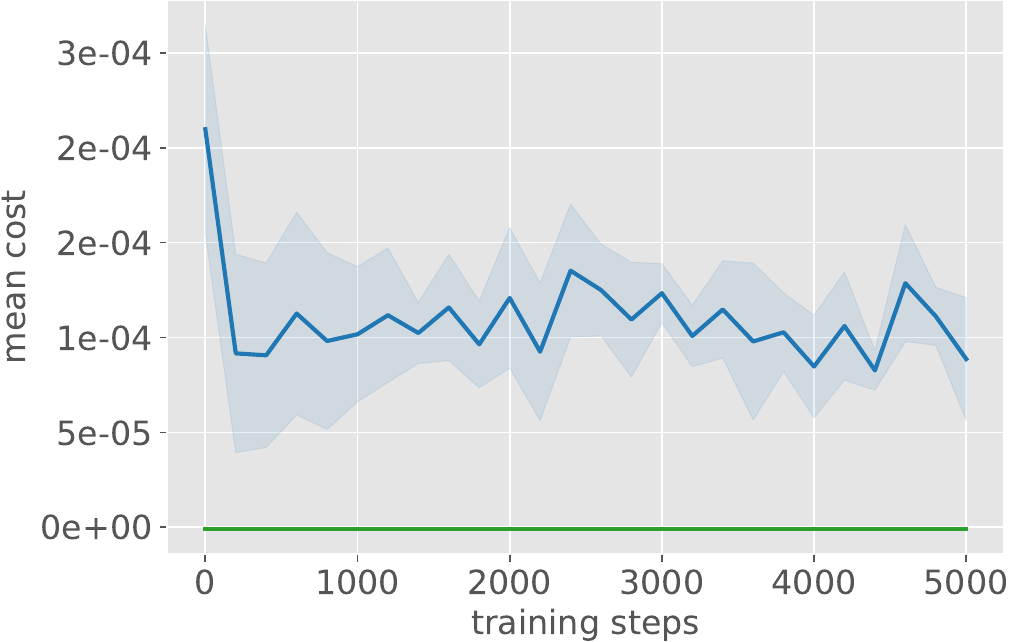} 
\caption{Mean violation.}
\label{fig:hcocost}
\end{subfigure}
% \begin{subfigure}[b]{1.0\columnwidth}
% \centering  
% \includegraphics[width=0.9\textwidth]{edis_tau.pdf} 
% \caption{Solutions distance.}
% \label{fig:hcovio}
% \end{subfigure}
\begin{subfigure}[b]{0.3\columnwidth}
\centering
\includegraphics[width=1.0\textwidth]{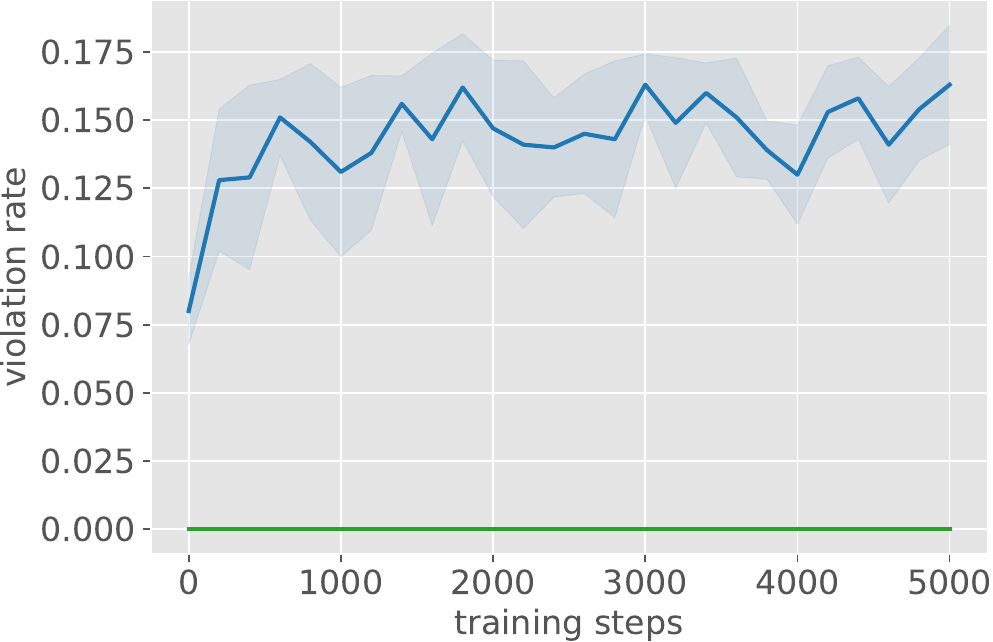} 
\caption{Violation rate.}
\label{fig:hcovio}
\end{subfigure}%
\caption{MSE loss, mean violation and violation rate for robotics tasks. The first row is for R+O03 and the second row is for HC+O. Mean violation is computed over violated samples and violation rate is the ratio of violated samples to all samples in the testing set. The shaded area for all the curves stands for standard deviation over 5 trials. Lower values are better for all the metrics.}\label{fig:rob}
\end{figure*}

It can be seen from \Cref{fig:edis} that with the decrease of temperature the accuracy of solutions obtained by DFWLayer becomes higher, as shown in \Cref{thm:gap}. However, the automatic differentiation involving a small temperature can be unstable after a large number of iterations. As presented in \Cref{fig:cos}, DFWLayer fails to compute accurate gradients when the temperature decreases to $\tau=0.25$ and $\tau=0.125$. Therefore, we choose to use an annealing temperature so that our method can obtain high-quality solutions and gradients at the same time, which is validated by the curve "$\tau$-as" (the purple lines) in \Cref{fig:tmp}. 

\subsubsection{Robotics Tasks Under Imitation Learning}
\label{sec:rob}

As for the field of robotics, in this subsection, we discuss the application of our method to specific tasks under an imitation learning framework. HC+O and R+O03 are constrained HalfCheetah and Reacher environments chosen from action-constrained-RL-benchmark \footnote{The benchmark can be accessed through \url{http://github.com/omron-sinicx/actionconstrained-RL-benchmark}} in \citet{kasaura2023benchmarking}. The original enviroments are from OpenAI Gym \citep{1606.01540gym} and PyBullet-Gym \citep{benelot2018pybulletgym} and the constraints are presented as follows:
\begin{align}
    \sum_{i}^{d} \lvert w_i a_i \rvert \le p_{\rm max},\label{eq:robcons}
\end{align}
where $w_i$ and $a_i$ are the angular velocity and the torques corresponding to $d$ joints, respectively, and $p_{\rm max}$ stands for the power constraint. Specifically, $d=6$ and $p_{\rm max}=20.0$ are for HC+O, while $d=2$ and $p_{\rm max}=0.3$ are for R+O03.

Under an imitation learning framework, optimization layers are added to the neural networks as the last layer, which aims to imitate expert policy and satisfy the power constraints \Cref{eq:robcons}. We first collected 1 million and 0.3 million expert demonstrations by running DPro, a variant of TD3 in \citet{kasaura2023benchmarking}, for HC+O and R+O03 respectively. Then, imitation learning was implemented using 80\% data as the training set with the MSE loss function. During the training phase, we tested the loss, mean violation and violation rate every 200 steps using the rest 20\% data as the testing set. 

The architecture for previous layers is $[400, 300]$ and the activation function is ${\rm ReLU}$ function. The parameters are updated by Adams optimizer with $1e-4$ as the learning rate. As for batch size, we have some experiments from \{8, 16, 32, 64, 128\}, and choose to use 64 considering the performance and efficiency comprehensively.

% \begin{figure}[ht]
% \centering

% \includegraphics[width=0.8\columnwidth]{CameraReady/LaTeX/figs/ro03_time.pdf} 
% \caption{Training time of a batch of 64 samples for R+O03. The error bars are made with standard deviation over 5 trials.}\label{fig:times}
% \end{figure}

As is shown in \Cref{fig:rob}, DFWLayer has significantly lower mean violation and violation rate than the other two baselines with comparable MSE loss. For R+O03 whose maximum power $p_{\rm max}=0.3$, the magnitude of the mean violation (about 3e-2) for Alt-Diff would have negative influence on the robotics tasks. However, DFWLayer and CvxpyLayer (with minor violation) are able to give feasible solutions to accomplish the tasks with accuracy and safety. For HC+O whose maximum power $p_{\rm max}=20.0$, CvxpyLayer achieves the best MSE loss with minor violation and a relatively high violation rate, while Alt-Diff satisfies the constraints all the time with the worst MSE loss. By contrast, DFWLayer outperforms the two methods comprehesively.

\end{document}